\documentclass{article}

\usepackage[normalem]{ulem}
\usepackage{booktabs}
\usepackage{tabularx}
\usepackage{multirow}
\usepackage{caption}
\usepackage{enumitem}
\usepackage{xfrac}
\usepackage{xcolor}
\usepackage{subfig}
\usepackage{color}
\usepackage{soul}
\usepackage{comment}
\usepackage{psfrag}
\usepackage{graphicx,epstopdf}
\usepackage{url}
\usepackage{adjustbox}
\usepackage{textcomp}

\usepackage{microtype}
\usepackage{hyperref}

\usepackage[accepted]{sysml2019}

\sysmltitlerunning{Restructuring Batch Normalization to Accelerate CNN Training}

\begin{document}

\twocolumn[
\sysmltitle{Restructuring Batch Normalization to Accelerate CNN Training}

\sysmlsetsymbol{equal}{*}

\begin{sysmlauthorlist}
  \sysmlauthor{Wonkyung Jung}{equal,snu}
  \sysmlauthor{Daejin Jung}{equal,samsung}
  \sysmlauthor{Byeongho Kim}{snu}
  \sysmlauthor{Sunjung Lee}{snu}
  \sysmlauthor{Wonjong Rhee}{snu}
  \sysmlauthor{Jung Ho Ahn}{snu}
\end{sysmlauthorlist}

\sysmlaffiliation{snu}{Department of Transdisciplinary Studies, Seoul National University, Seoul, Republic of Korea}
\sysmlaffiliation{samsung}{Samsung Electronics, Suwon, Republic of Korea}

\sysmlcorrespondingauthor{Jung Ho Ahn}{gajh@snu.ac.kr}
\sysmlcorrespondingauthor{Wonjong Rhee}{wrhee@snu.ac.kr}

\sysmlkeywords{Convolutional Neural Networks, Batch Normalization, Kernel Fusion, Kernel Fission, Kernel Restructuring}

\vskip 0.3in

\begin{abstract}
  %
  %

  %

  Batch Normalization (BN) has become a core design block of modern Convolutional
  Neural Networks (CNNs). 
  %
  A typical modern CNN has a large number
  of BN layers in its lean and deep architecture.
  BN requires mean and variance calculations over 
  each mini-batch during training.
  Therefore, the existing memory access reduction techniques, such as
  fusing multiple CONV layers, are not effective for accelerating 
  BN due to their inability to optimize mini-batch related calculations during training.
  To address this increasingly important problem, we propose to 
  restructure BN layers by first splitting a BN layer into two sub-layers (fission) and
  then combining the first sub-layer with its preceding CONV layer and 
  the second sub-layer with the following activation and CONV layers
  (fusion).
  The proposed solution can significantly reduce main-memory accesses while training 
  the latest CNN models, and the experiments on a chip multiprocessor
  show that the proposed BN restructuring can improve
  the performance of DenseNet-121
  by 25.7\%.
\end{abstract}
]

\printAffiliationsAndNotice{\sysmlEqualContribution} 


\section{Introduction}
\label{sec:introduction}
Deep Neural Networks (DNNs) have emerged as the key technique of artificial intelligence, and
Convolutional Neural Networks (CNNs) 
are widely used for image classification and object detection tasks.
Typical DNNs, which consist of multiple stacked layers and multiple channels (filters) per layer, require
billions of operations~\cite{arxiv-2016-dnn-analysis}
for training and inference, and CNNs usually require larger amount of operations than other types of DNNs.
As the first order of solution to the computational load problem, hardware accelerators
such as GPGPUs~\cite{ieeemicro-2017-nvidia-pascal}, FPGAs~\cite{micro-2016-fused}, ASICs~\cite{isca-2017-tpu},
and manycore processors~\cite{ieeemicro-2016-knl} have been developed.
Then, further optimizations have been performed specifically for CNNs.
For the CNNs designed in the early days, convolutional (CONV) and fully-connected (FC) layers
were the most time-consuming parts, and the CNN accelerator research
has mainly focused on optimizing the two layer types.
For example, loop blocking and reordering techniques were 
shown to be highly 
effective for maximizing data reuse of CONV/FC layers~\cite{arxiv-2016-systematic, isca-2016-eyeriss},
and subsequently network compression or approximate computing methods were introduced to significantly
reduce computation and memory access~\cite{isca-2016-eie, iclr-2015-deepcompression, nips-2015-prunning, arxiv-2015-datafree}.
Previous works, however, paid almost no attention to the \emph{non-CONV} layers
as their influence on resource consumption was ignorable.\footnote{We address
layers other than CONV/FC as non-CONV layers.} 

The latest CNN models such as ResNet~\cite{cvpr-2016-resnet} and DenseNet~\cite{cvpr-2017-densenet}, however, 
are actively adopting new
structures and non-CONV layers to improve prediction performance.
For instance, \emph{skip connection} has been utilized to stabilize backpropagation and to
enable stacking hundreds of layers;
and Batch Normalization (BN) has been developed with the original goal of addressing \emph{internal covariate shift}
phenomenon~\cite{pmlr-2015-bn}.  
With the invention of the new layer types that enables more layers to be
stacked, the relative portion of non-CONV layers in a CNN model has been increasing. 
%
%
%
%
On the contrary, the computation load of CONV layers has been declining
due to a reduction in the size of the convolution filters.
The early CNN models such as AlexNet~\cite{nips-2012-alexnet} and 
VGG~\cite{arxiv-2014-vggnet} have used convolution filters with 
the size of 3$\times$3, 5$\times$5, and even 11$\times$11. 
But the recent designs (e.g., ResNet and DenseNet) apply 1$\times$1 or
3$\times$3 filters 
in most CONV layers and successfully reduce the computational overhead.
Overall, the design trends of modern CNNs indicate that the importance of CONV/FC layers
is decreasing whereas the importance of non-CONV layers is increasing.

While we expect many non-CONV layers to be developed and become a significant part
of computational load, BN during the training is known as 
the most computationally intensive non-CONV layer as of today.
%
%
One might think inference is much more important because training needs to be completed only once; 
but in reality, training needs to be performed repeatedly by trying different hyperparameters such as
network depth, number of neurons, learning rate, regularizer, optimizer, and activation
function~\cite{bergstra2012random, snoek2012practical}.
Even after a training is deemed to be complete and the trained network is deployed for a real
service, deep learning research is moving so fast that the developers are immediately forced to consider
newly invented solutions to replace the deployed one.
Therefore, both researchers and practitioners end up spending a significant portion of
their computational resources on training, and improving the performance of training is as important as
(or arguably even more important than) improving that of inference.
%

During training, a BN layer requires per-channel mean and variance of input elements to be evaluated
over the entire mini-batch dataset.
There are other layers and operations that require statistics to be calculated over
the entire mini-batch dataset (e.g., covariance matrix for regularization~\cite{cogswell2015reducing} and
mutual information for disentangling GAN representations~\cite{chen2016infogan}),
but BN has the largest impact on computational load as it is currently very popular
and it can be included in multiple places within a deep learning network.
%
%
As in \cite{micro-2016-fused} where intermediate data between layers was recognized as
an opportunity to accelerate CNN inference, we recognize that the mini-batch calculation
of BN is an opportunity as well. 
However, as opposed to \cite{micro-2016-fused}, we focus on restructuring BN
because BN during training imposes strict dependency across a large volume of mini-batch dataset
which does not fit within on-chip buffers and hence fusing multiple convolutional
layers is less attractive.

In this paper, we develop a BN restructuring solution for the latest CNN models
with the following key contributions:
%
\begin{itemize}
\item We explore the execution time breakdowns and show that non-CONV layers have become
significantly more important for the latest CNN models.
We show that BN is the most important among non-CONV layer types.
\item We propose a novel BN-layer restructuring solution where a BN layer is first divided 
  into sub-layers (fission) and then merged with neighboring layers (fusion).
This restructuring can significantly mitigate memory bottleneck problem by reducing memory
traffic concentrated on BN layers.
\item We achieve 25.7\% (16.1\%) of performance enhancement on the latest chip multiprocessor 
  (Intel's Skylake~\cite{skylake}) on top of 
  a highly-optimized CNN library~\cite{mkldnn} for DenseNet-121 (ResNet-50).
    Applying the BN restructuring to GPU with an open-source
    linear algebra library~\cite{nvidia-cutlass} also improves
    performance of DenseNet-121  and ResNet-50 by 17.5\% and 7.8\%, 
    respectively.\footnote{Check our implementation at 
    \url{https://github.com/scale-snu/caffe-bn-restructuring} and
    \url{https://github.com/scale-snu/mkldnn-bn-restructuring}. }
%
%
\end{itemize}

\section{Background and Motivation}
\label{sec:background}

\begin{figure}[!tb]
  \center
  \includegraphics[width=\columnwidth]{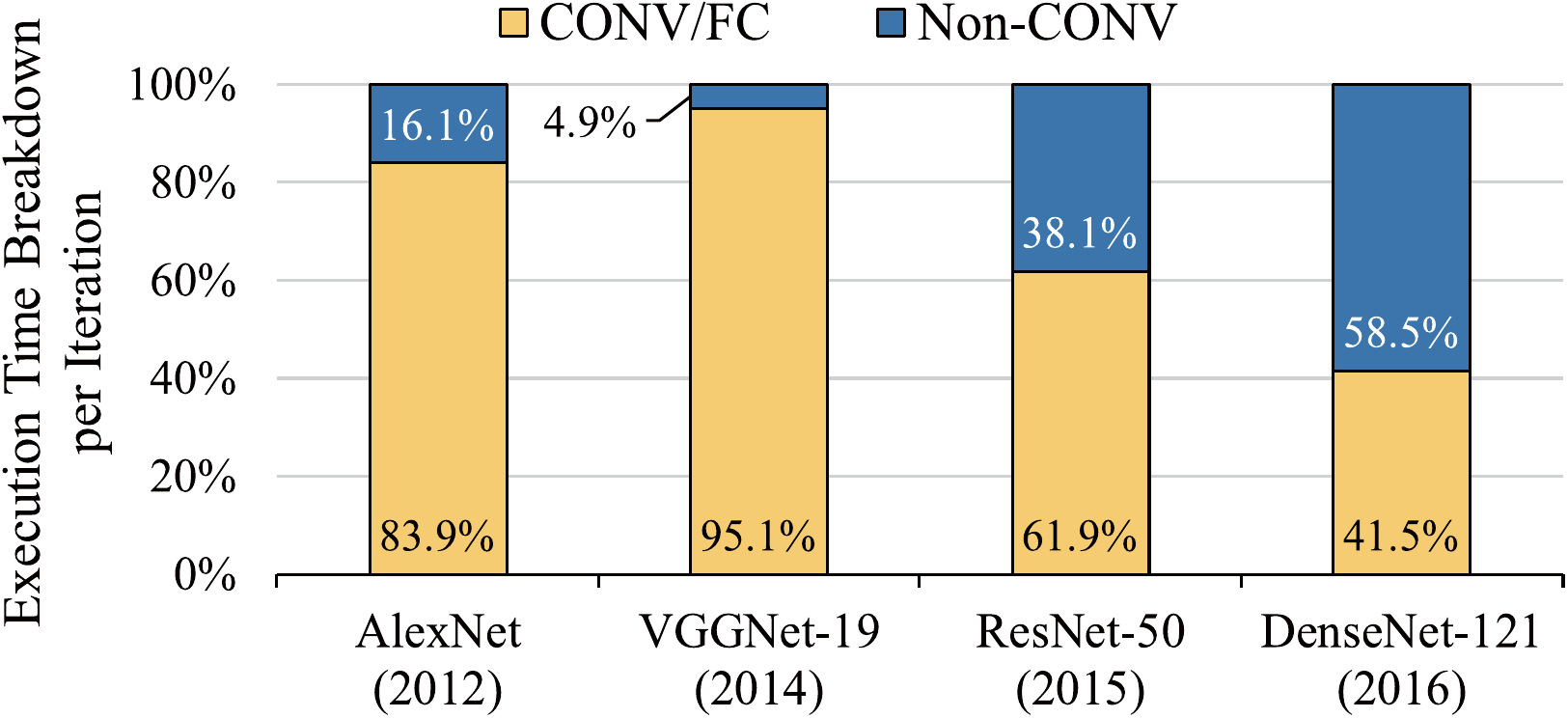}
  \caption{Execution time breakdown of popular CNN models over layer
  types.
  We categorize CONV and FC layers into a group (CONV/FC) and the 
  remaining layers into the other (non-CONV).
  CONV/FC layers dominate the execution time for the early models, but
  non-CONV layers are also important for the later models that are much deeper.}
  \label{fig:time-portion-over-CNNs}
  \vspace{-0.1in}
\end{figure}

\subsection{Trends in CNN accelerator designs}
\label{sec:background-CNN-accelerators}
Many CNN accelerator proposals and designs are mainly focused on
convolutional (CONV) and fully-connected (FC) layers.
This is because those layers take up most of the inference and
training time of relatively shallow CNNs (up to few dozens of layers).
For example, AlexNet~\cite{nips-2012-alexnet} consists of only 5 CONV layers
and 3 FC layers, whereas VGGNet~\cite{arxiv-2014-vggnet}
has 13-16 CONV layers and 3 FC layers.
On these early and shallow CNN models, the portion of execution time 
on CONV and FC layers is dominant, accounting for up to 95\% of total
execution time as shown in Figure~\ref{fig:time-portion-over-CNNs}
(we measured the training time from the system specified in 
Section~\ref{sec:experimental-setup}).

Common optimization strategies implemented in the CNN accelerators for
the two types of layers include: reducing memory bandwidth by
maximizing the data reuse of weights, input feature maps, and output
feature maps~\cite{isca-2015-shidiannao,
isca-2016-eyeriss}, pruning redundant parameters (e.g., weights) and
exploiting sparsity~\cite{isca-2016-cnvlutin, isca-2016-eie}, computing in an approximate 
manner~\cite{iclr-2015-deepcompression, nips-2015-prunning}, and
adopting new memory technologies~\cite{isca-2016-prime,
isca-2016-neurocube, isca-2016-isaac}.
However, there are relatively few studies~\cite{isca-2018-gist}
focusing on optimizing the
layers other than CONV/FC layers (non-CONV), such as ReLU
(Rectified Linear Unit),
pooling, and batch normalization (BN) layers because \emph{these 
non-CONV layers take up much less time compared to the CONV/FC
layers on these shallow CNNs.}

Recent deeper CNN models (e.g., ResNet~\cite{cvpr-2016-resnet} and
DenseNet~\cite{cvpr-2017-densenet}) spend a larger portion of 
time executing non-CONV layers as opposed to the conventional shallow CNNs.
%
For example, DenseNet-121 (DenseNet with 120 CONV layers plus one FC layer)
spends more than half of the execution time on the non-CONV layers 
(see Figure~\ref{fig:time-portion-over-CNNs}).
For these CNN models, accelerating the non-CONV layers is becoming increasingly
important.

Optimizing the non-CONV layers is more complicated when it comes to 
training compared to inference.
Previous techniques to boost the inference process, such
as approximate computing and compression, are not easily applicable to 
training as weights are updated in the course of
training.
Complex data dependencies during training also matter. A layer like BN 
needs only element-wise operations during inference, 
but it demands data of multiple intermediate output values from its previous layers
during training, making the optimization a non-trivial task.

The training process matters in that it requires significant computing
costs.
ResNet-50 takes 29 hours to train with 8 Tesla P100 GPUs, as each epoch
(training an entire dataset) consumes 16 minutes for the images of the
the ImageNet Large Scale Visual 
Recognition Challenge dataset~\cite{arxiv-2017-facebook}.
Google AlphaGo~\cite{nature-2016-alphago} was trained for more than 
three weeks with 50 GPUs to beat a top-class professional Go player.
If training is an one-time event, its cost would be well amortized.
As DNN models evolve rapidly and more data are accumulated, however,
frequent or even continuous training is needed.
These all support the importance of optimizing the training
process.

So far, however, relatively few studies have focused on training,
especially the non-CONV layers.
Although several DNN acceleration strategies have been studied to
mitigate the computing cost of training~\cite{arxiv-2017-facebook,
micro-2016-vdnn, isca-2017-scaledeep}, these are different from our work in that they
did not focus on accelerating the non-CONV layers.
%

\begin{figure*}[!tb]
  \center
  \includegraphics[width=\textwidth]{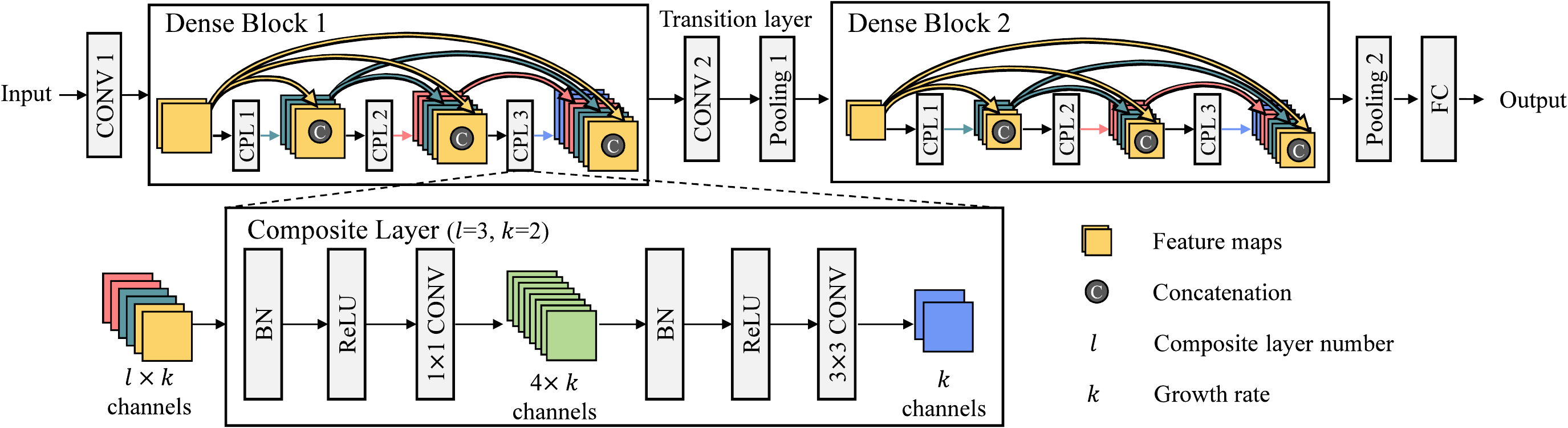}
  \caption{
    An exemplar DensetNet structure with two Dense Blocks, connected through a transition layer which changes the number and size of the input feature maps to a Dense Block.
    Each Dense Block has multiple Composite Layers (CPLs), each of which is connected to every other CPL within a Dense Block
    in a feed-forward fashion.
    A CPL consists of six layers (BN, ReLU, 1$\times$1 CONV, BN, ReLU, and 3$\times$3 CONV).
    The 1$\times$1 CONV layer in a CPL, called bottleneck layer, limits the number of input feature maps to 4$\times$\textsf{k} while the second CONV outputs \textsf{k} channels that are concatenated
    to the input feature maps.
    Growth rate (\textsf{k}) is the variable for how many feature maps are concatenated per CPL; feature maps stack up as they go through CPLs.
  }
  \label{fig:densenet-structure}
\end{figure*}

\vspace{-0.05in}
\subsection{Deep CNN models}
\label{subsec:recent-CNNs}
%
%
%
Deep networks are known to have better expressivity and better optimization characteristics, 
and deep networks are widely used with convolutional layers. To train a deep CNN model, stable propagations of 
activations in the forward path and gradients in the backward path are crucial. Guaranteeing 
a stable training, however, is still an incompletely understood problem. 

BN~\cite{pmlr-2015-bn} was proposed in 2015 for stable learning,
where it stabilizes the distribution of nonlinear inputs 
by normalizing the samples within a mini-batch. 
Specifically, BN first computes the mean and variance values for
each channel of its input feature maps sweeping through the values
over a mini-batch; then it performs a normalization with
a scaling factor ($\gamma$) and a shift factor ($\beta$).
%
%
While BN's stabilizing and regularizing effects have been confirmed through 
numerous empirical studies that followed the original study, the reason why BN works 
well has been controversial. Recent works shed some light on the reason, where \cite{nips-2018-bnhelpoptimize} show that internal covariate shift has 
little to do with the success of BN but that BN makes the optimization landscape significantly smoother and  
\cite{nips-2018-understandingbn} demonstrate how large gradient 
updates can result in diverging loss and activations growing uncontrollably with network depth and how BN avoids these. 
The recent explanations are in line with the popular design of heavily utilizing BN in deep CNN models.

Residual learning is known to be helpful for training deep CNN networks, too. 
%
%
Residual learning uses \emph{skip connection} where 
a layer can skip some layers and connect directly
to a farther-away layer as well as its adjacent layer
(through a layer called Split).
ResNet, one of the state-of-the-art CNNs, adopted
residual learning with identity mapping, which adds a layer to the far
layer in an element-wise manner (through an element-wise sum (EWS) layer).
This effectively shortens the distance between close-to-input layers
and close-to-output layers, helping very deep networks with 
hundreds of layers to converge.
Due to its record-breaking performance in image recognition,
residual learning has been frequently adopted in the latest 
CNNs~\cite{cvpr-2016-resnet, arxiv-2016-identity, arxiv-2016-resnext}.
DenseNet, which we mainly 
target for optimization in this work, 
is also a variant of the residual learning with 
a slight but important difference -- it concatenates multiple feature maps
rather than performing EWS, through a layer called Concat. 

%

The importance of optimizing these non-CONV layers is growing as they continue to take
a larger portion of the training time.
Because smaller CONV filters are more frequently used on modern 
deeper CNNs, the amount of computation per CONV filter has been
decreasing.
For example, AlexNet utilizes 11$\times$11 and 5$\times$5 filters, whereas more recent models 
such as DenseNet, ResNet, and Inception network~\cite{arxiv-2016-inception-v4},
adopt relatively small filters, mainly 3$\times$3 and 1$\times$1.

\vspace{-0.05in}
\subsection{DenseNet: a state-of-the-art CNN model}
\label{sec:background-densenet}
DenseNet~\cite{cvpr-2017-densenet} is an example of more recent
CNN models that use both BN and residual learning to achieve 
high accuracy with deeper (possibly surpassing 100) layers and
smaller CONV filters.
It achieves classification accuracy comparable to ResNet, but
with fewer learning parameters.
A key feature of DenseNet lies in its dense connectivity.
A DenseNet is a sequence of \emph{Dense Blocks} (see
Figure~\ref{fig:densenet-structure}); two adjacent
Dense Blocks are connected through transition layers (e.g., few 
CONV and pooling layers).
Each Dense Block has multiple \emph{Composite Layers} (CPLs).
A CPL consists of a sequence of BN, ReLU, 1$\times$1 CONV,
BN, ReLU, and 3$\times$3 CONV layers.
The \textsf{l}'th CPL within a Dense Block receives \textsf{l}$\times$\textsf{k} input
feature maps (channels).
The first 1$\times$1 CONV layer limits the number of feature
maps to \textsf{m}$\times$\textsf{k}; when \textsf{l} $>$ \textsf{m}, the 1$\times$1 CONV layer effectively reduces the
computation cost of its following sequence of BN, ReLU, and
$3\times3$ CONV layers, and hence it is called a bottleneck layer.
All CPLs within a Dense Block is fully connected utilizing Concat
and Split layers in a feed-forward fashion (connections not forming
a cycle).
Feature map size grows within a Dense Block;
because the output feature maps from the preceding CPLs are 
concatenated, a CPL located later (farther from input) in a Dense
Block has more input feature maps (channel).
A transition layer works similar to a bottleneck layer, limiting 
the number of channels.

Dense connectivity enables DenseNet to reduce computation cost
from CONV layers.
It embodies residual learning exploiting skip connection;
however, as opposed to ResNet which performs EWS, DenseNet 
concatenates feature maps from preceding layers.
%
%
CPL exploits the observation from empirical studies where
placing the BN layer before ReLU and CONV layers provided better
recognition performance~\cite{cvpr-2017-densenet}.
Furthermore, DenseNet has shown that relatively small $k$ is
sufficient for achieving a state-of-the-art accuracy on recognition,
as information in feature maps is transferred through dense
connectivity.  
The result is higher computation efficiency in these CONV layers.
While non-CONV layers such as BN, ReLU, and Split have been gaining
prominence not only in DenseNet but also in other modern
CNNs~\cite{cvpr-2016-resnet, arxiv-2017-mobilenet, arxiv-2016-inception-v4},
an improved degree of computation efficiency in CONV layers due to
the aforementioned reasons makes optimizations for non-CONV layers
even more important.
In this work, we examine the characteristics of computational challenges
in DenseNet and propose solutions to overcome these.

\begin{figure*}[!tb]
  \center
  \includegraphics[width=\textwidth]{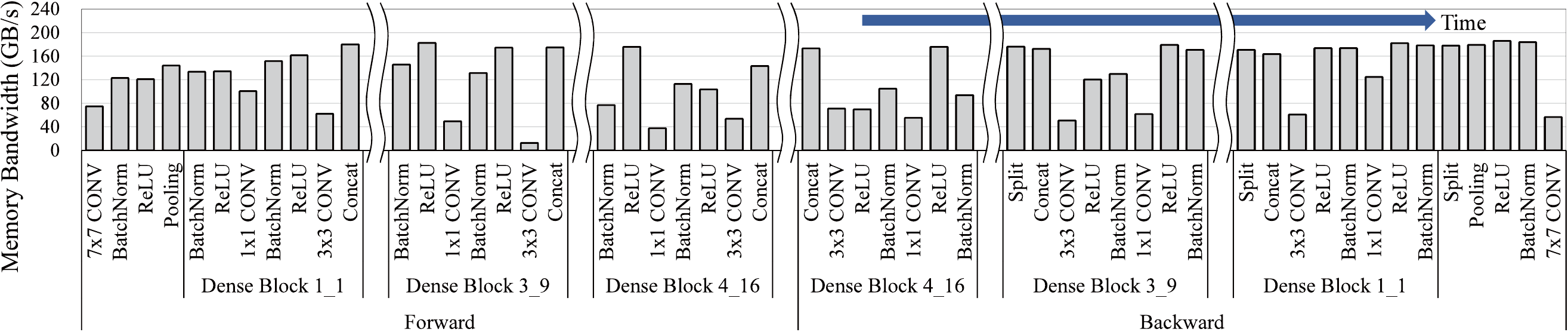}
  \vspace{-0.08in}
  \caption{
    Memory bandwidth utilization of layers on the DenseNet-121 CNN
    model over time~\cite{cvpr-2017-densenet}.
    Each CONV layer is interleaved with other operation types, such as
    BN, ReLU, and Concat (concatenation), each exhibiting different
    degrees of bandwidth demand.
    Peak main-memory bandwidth of the evaluated system is 230.4GB/s
    (12 DDR4-2400 channels).
  }
  \label{fig:bw-over-time}
\end{figure*}

\section{BN Restructuring}
\label{sec:contribution}

\subsection{Analyzing DenseNet}
\label{sec:contribution-1}
In this section, we mainly focus on analyzing the memory access characteristics
over time for training 
DenseNet-121. 
(ResNet is also considered for the final evaluation in Section~\ref{sec:evaluation}.)
%
For the analysis, we use the latest chip multiprocessor (Intel
Skylake~\cite{skylake}) that can provide decent memory bandwidth and computing power
with the capability to measure various architectural statistics (e.g., memory accesses
and cache misses).
%
%

We observed the following key memory access characteristics when
training DenseNet.
First, similar to other CNN models, DenseNet exhibits
repeating patterns in memory traffic because layers are processed
sequentially and a set of layers with 
different computational characteristics are executed repeatedly.
Second, memory bandwidth demands are different from layer to layer
even for the same layer type.
This is because each layer's
number of input feature maps (ifmaps), number of output feature maps (ofmaps), and
filter size can be different for the purpose of recognition performance and efficient operation.
For example, CONV layers with smaller filter sizes have relatively
high memory access rates compared to the total amount of 
computation because the reuse rate of ifmaps within on-chip buffers
is low, whereas layers with relatively large filter sizes can
reduce the off-chip memory bandwidth demand due to the higher reuse rate of
ifmaps.
Actually, DenseNet has higher memory bandwidth demands compared to
early CNN models as it uses 1$\times$1 CONV and 3$\times$3 CONV
layers except the first 7$\times$7 CONV layer (see 
Figure~\ref{fig:bw-over-time}).

Memory bandwidth demand varies substantially between CONV layers and 
non-CONV layers due to the difference in computational 
characteristics.
The non-CONV layers of DenseNet-121 are mostly bottlenecked by the
peak main-memory bandwidth of the system we use 
(230.4GB/s),
whereas the CONV layers 
underutilize the available bandwidth (only up to 120GB/s).
This is because the non-CONV layers have less data locality and 
computation intensity, which makes loop blocking techniques 
less effective, leading to higher demand in memory bandwidth. 
%

\begin{figure}[!tb]
  \center
  \includegraphics[width=2.5in]{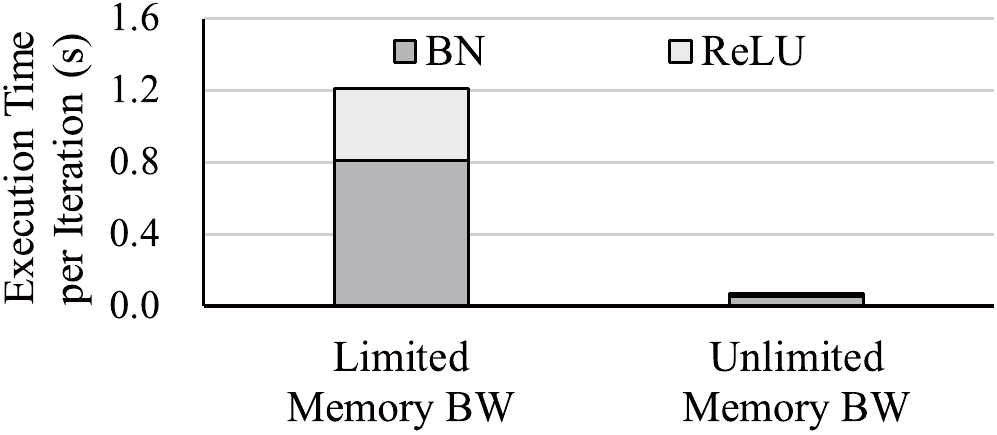}
  \caption{
    Comparing execution time of BN and ReLU layers
    with finite and infinite (hypothetical) 
    memory bandwidth.}
  \label{fig:unlimited-bw}
  \vspace{-0.04in}
\end{figure}

If a system has infinite memory bandwidth, the high memory bandwidth
demand of the non-CONV layers would not be a problem as the execution
time of them would be limited by computation power (peak FLOPS) of the
system and the non-CONV layers are far less computationally intensive
compared to the CONV layers. 
However, improving the performance (i.e., bandwidth and latency) of
the main-memory systems has remained relatively challenging, as
opposed to the rapid growth in computational power (i.e., FLOPS or
OPS) per device.
Modern data-parallel architectures used for CNN acceleration
still provide around hundreds of GB/s of main-memory bandwidth (e.g., 732GB/s with 
Nvidia Tesla P100~\cite{ieeemicro-2017-nvidia-pascal}) due to the power and signal integrity issues;
even if the absolute data transfer rate is impressive, it is not high enough 
to completely satisfy the bandwidth demands from memory intensive 
(computationally less intensive) layers, especially BN and ReLU.
The maximum computing power of Tesla P100 is 10.6TFLOPS (single-precision
floating point), being translated to 14.5FLOP/B or 58 FLOPs per 32-bit
data.
Assuming that ReLU requires one clipping operation per 32-bit data, 58$\times$
more main-memory bandwidth is needed to match P100's computing power.

We tested the BN and ReLU layers of DenseNet using a real machine (Intel
Skylake) with a hypothetical model providing infinite memory bandwidth for 
the BN and ReLU layers (Figure~\ref{fig:unlimited-bw}).
To simulate the hypothetical machine, we made BN and ReLU layers skip DRAM
memory accesses by manipulating memory address offsets, thereby all data 
accesses can fit in L1 caches but the number of operations being executed
remains unchanged.
In this experiment, we exclude Concat and Split layers because such layers
mainly perform memory copies that can be optimized by passing pointers 
instead of physical copies even if they demand high memory bandwidth in
our reference implementation~\cite{densenet-github} using 
MKL-DNN~\cite{mkldnn}.
We observe that without the bandwidth limitation, the BN and ReLU layers
enjoy 20$\times$ speedup.
If computational power (peak FLOPS) is improved faster than main-memory
bandwidth (peak GB/s) in future CNN accelerators, which is quite likely
as computation is cheap and communication is 
expensive~\cite{isca-2016-eie} in contemporary VLSI systems,
these non-CONV layers could possess even larger portion of execution
time in future.
%
Therefore, it is important to reduce memory accesses on those layers.

Memory accesses in ReLU and BN layers mostly come from reading and
writing ifmaps and ofmaps.
At a given layer, the aggregate size of these feature maps across 
a mini-batch could be too large to fit in on-chip memory with MBs of 
capacity especially when the size of a mini-batch reaches or surpasses
100, which is preferred as larger mini-batch sizes typically lead to
better hardware utilization.

To decrease the memory accesses in these layers, we can consider
maximizing data reuse of the ifmaps and the ofmaps between and within
the layers.
\emph{However, because BN layers have strict data dependency, cross-layer
data reuse is prohibited.}
In the forward pass of a BN layer, all pixels of ifmaps that belong to
a mini-batch should be retrieved to get per-channel mean and variance
values prior to normalizing individual pixels.
Likewise, in backpropagation, calculating the partial derivatives on $\gamma$
(scaling factors) and $\beta$ (shift factors) accompanies sweeping
the partial derivatives on ofmaps; this should precede computing the partial
derivatives on ifmaps.
Because these dependencies make data reuse distance far in BN, it is
difficult to apply the data reuse techniques that were previously
proposed for CONV layers to these non-CONV layers.

%
%
A prior work~\cite{micro-2016-fused} fuses multiple CONV
layers to facilitate data reuse in-between; 
however, because BN
layers are frequently inserted between CONV layers (e.g., each
CONV layer within a CPL of DenseNet 
is paired with a BN layer) and
BN imposes strict dependency, inter-CONV fusion is not attractive to
the latest deeper CNNs.
With this motivation, we suggest \emph{Fission-n-Fusion}, a novel solution 
that copes with this strict data dependency in BN layers 
to minimize unnecessary data transfers.

\begin{figure*}[!tb]
  \center
  \includegraphics[width=\textwidth]{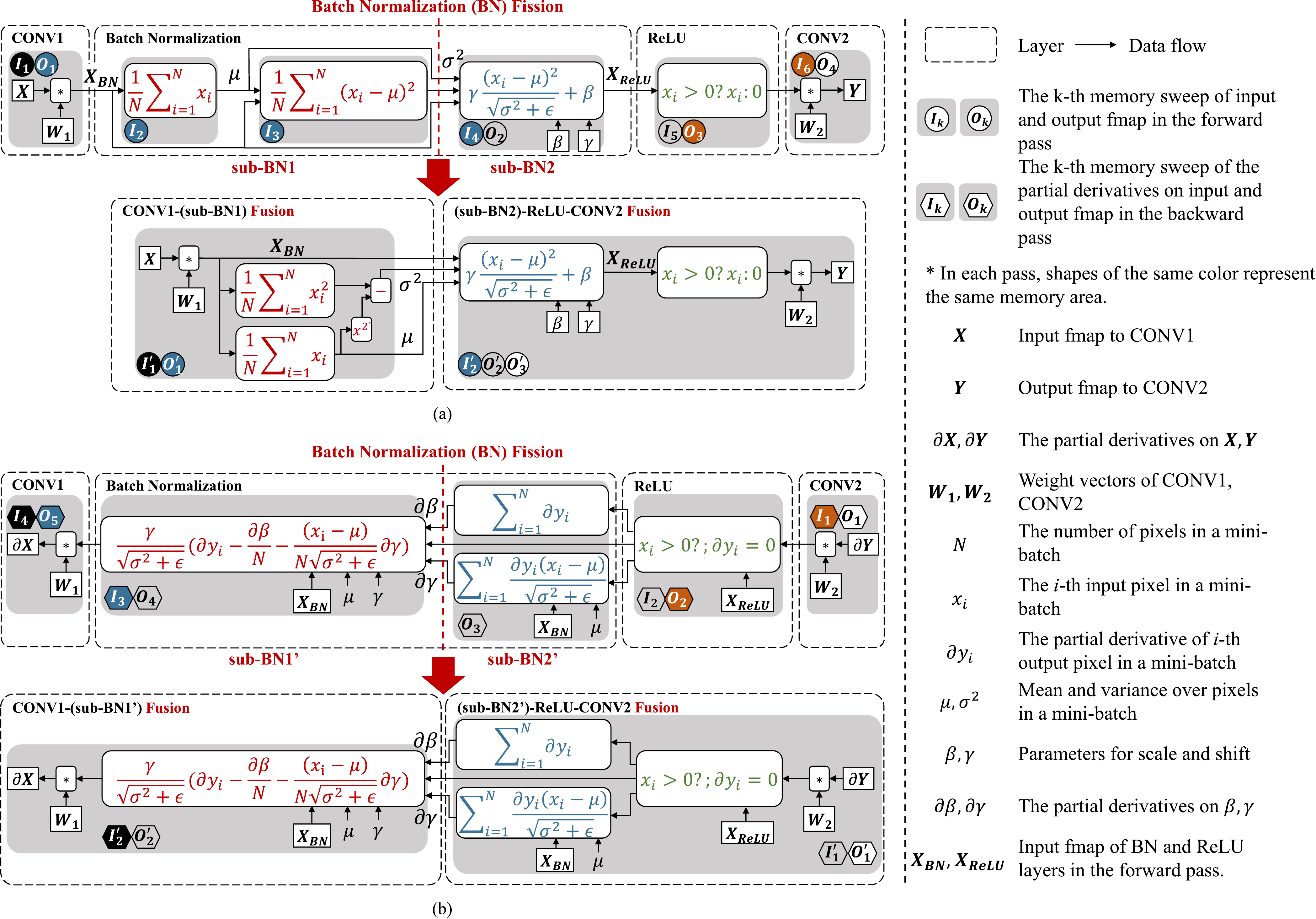}
  \caption{
    Illustration of BN Fission-n-Fusion on (a) the forward pass and
    (b) the backward pass. 
    We present data dependency with arrows, memory sweeps (either
    reading or writing 100+ feature maps across an entire mini-batch
    per channel) as grey boxes, and multiple memory accesses to the
    same feature map with shapes of the same color.
    A BN layer is first divided into two sub-layers and then fused
    with the preceding CONV and the following CONV (+ReLU) layers,
    respectively.
  }
  \label{fig:fission-and-fusion}
\end{figure*}

\subsection{Fission-n-Fusion}
\label{sec:contribution-2}

We first divide the BN layers within the CPLs of DenseNet
into two sub-layers (Fission). 
Then, we combine the first sub-layer with its preceding CONV layer and 
the second sub-layer with both the ReLU layer and the following CONV 
layer (Fusion). 
Fusing multiple (sub-)layers can significantly reduce off-chip memory
accesses of BN layers. 
Fission aids Fusion to be more effective by enabling the operations
of a BN layer with the strict data dependency to be entirely fused 
to its preceding and following CONV layer, through which all
memory accesses from the original BN layers can be 
removed.
The resulting combination of the two ideas, BN Fission-n-Fusion 
(BNFF), can save a significant amount of memory access per training
step, which is translated to substantial performance improvement. 

Figure~\ref{fig:fission-and-fusion} illustrates how our proposed BNFF
is applied to the reference DenseNet implementation~\cite{densenet-github}
that is represented using computational graphs.
A grey box is a computational node that accompanies sweeping
(reading or writing) all ifmaps or ofmaps of a channel
within a mini-batch; this sweeping cannot be filtered by on-chip
buffers because the aggregate size of an ifmap and ofmap
across a mini-batch of 100 or more images is too large to fit in
on-chip buffers, as explained in Section~\ref{sec:contribution-1}.
An arrow indicates data flow between the computational node, composing
dependency.
A dotted box represents a layer before and after BNFF.
We also present multiple memory accesses to the same feature map with
the shapes of the same color (e.g., O$_\texttt{1}$, I$_\texttt{2}$,
I$_\texttt{3}$, and I$_\texttt{4}$ in the forward pass).

During training, a BN layer of the reference implementation accesses
the same ifmap and the partial derivatives on the ofmap multiple times 
due to data dependency on both forward and backward passes.
On the forward pass, a BN layer reads ifmaps three times 
(I$_\texttt{2}$, I$_\texttt{3}$, I$_\texttt{4}$ in the upper half of
Figure~\ref{fig:fission-and-fusion}(a)) 
and writes its results to ofmaps once (O$_\texttt{2}$).
Here, the first and the second read of ifmaps are for computing
the mean ($\mu$) and variance ($\sigma^2$) of pixels through an entire
mini-batch per channel, whereas the last read is for normalization. 
These multiple ifmap reads are due to the dependency between the 
computational nodes within a BN layer (i.e., computing variance 
needs mean, and normalization needs mean and variance). 
Similarly, on the backward pass, data dependency exists but in a 
reverse direction. 
The partial derivatives on the ofmap from the ReLU layer
are read multiple times to compute the partial derivatives on $\gamma$ and
$\beta$, which are needed for the partial derivatives on the ifmap.

We separate the normalization operations from mean and variance computation
in a BN layer, calling the divided layers as sub-BN1 and sub-BN2,
as depicted in the lower half of Figure~\ref{fig:fission-and-fusion}(a). 
Fission enables both sub-BNx layers to be fused with its adjacent layers; 
sub-BN1 glues to CONV1 (the CONV layer that precedes the BN layer being
split) whereas sub-BN2 glues to ReLU and CONV2 (the CONV layer that follows
the BN layer in Figure~\ref{fig:fission-and-fusion}). 
Fission provides an opportunity to handle operations of each sub-BNx during
reading and writing ifmaps/ofmaps in BN's preceding and following CONV
layers without additional off-chip memory access.
Without Fission, a BN layer can be fused with either CONV1 or ReLU-CONV2 only,
which prevents the maximum reduction of memory accesses by Fusion.

For the two aforementioned dependencies within a BN layer (computing variance
needs mean, and normalization needs mean and variance), Fission copes with
the latter, but the former still exists, especially on the forward pass.
We eliminate this dependency by exploiting a simple mathematical formula below:
\begin{center}
  \vspace{-0.02in}
${V(X) = E((X{-}E(X))^2) = E(X^2) - E(X)^2}$
  \vspace{-0.02in}
\end{center}
That is, we can compute the per-channel variance of the pixels across a 
mini-batch of BN's ifmaps by computing $E(X^2)$ together with $E(X)$ from
a single memory sweep.
This mean/variance fusion (MVF) enables the two memory sweeps 
(I$_\texttt{2}$, I$_\texttt{3}$) within sub-BN1 to be merged
with O$_\texttt{1}$ in CONV1 through Fusion.
Computing the expectation of the square ($E(X^2)$) is more susceptible to
accuracy errors which are inherent in floating-point arithmetic compared
to computing $E(X)$.
If it affects the accuracy of the network, we can use higher-precision
representations (e.g., double-precision) to
store intermediate data. 
Because BN is limited by main-memory bandwidth even after 
applying BNFF, using higher-precision representations and arithmetic does
not impact training performance.
During experiments, we observed that using single-precision floating-point
arithmetic is good enough for calculating $E(X^2)$.

To fuse the first sub-BN1 layer with the CONV1 layer, we
simultaneously perform both operations: convolution of CONV1 and
accumulating 
$x_\texttt{i}$ and ${x_\texttt{i}}^2$ from all pixels of CONV1's
ofmaps over a mini-batch of BN on each channel.
This ends up making one fused layer, CONV1-(sub-BN1), removing
all memory accesses of sub-BN1.
The second sub-BN2 layer is fused with both the ReLU layer and
the CONV2 layer; 
the fused (sub-BN2)-ReLU-CONV2 layer perform normalization, ReLU,
and convolution all together.

Because ReLU performs element-wise clipping, the recent CNN 
library we use~\cite{mkldnn} fuses a ReLU layer to its 
preceding CONV layer in the process of creating the ofmaps
of the CONV layer.
However, the latest CNN models might put ReLU layers in a different order.
For example, in DenseNet, a ReLU layer is placed prior to the
following CONV layer, and hence the aforementioned fusion 
implementation does not work.
Our ReLU-CONV fusion (RCF) implementation executes the ReLU operation
in the process of reading the ifmaps of the following CONV layer,
removing memory access by the ReLU layer.
%
Putting all the proposed solutions together, three memory sweeps
(O$_\texttt{1}$, I$_\texttt{2}$, I$_\texttt{3}$) are reduced into
one sweep (O$_\texttt{1}$') at the first fused layer, and five 
(I$_\texttt{4}$, I$_\texttt{5}$, I$_\texttt{6}$, O$_\texttt{2}$,
O$_\texttt{3}$) into two (I$_\texttt{2}$', O$_\texttt{2}$') 
at the second fused layer. 
A memory sweep (O$_\texttt{2}$') is required as it is used
once again during the BN operation in the backward pass.

In the backward pass (backpropagation), BN first calculates the partial derivatives
on $\gamma$ and $\beta$, and then
computes the partial derivatives on the ifmaps with them.
This forms a strict data dependency.
We separate operations that compute the partial derivatives on BN's ifmaps from
calculating the partial derivatives on $\gamma$ and $\beta$, 
calling the divided layers as sub-BN1' and sub-BN2', as depicted in 
Figure~\ref{fig:fission-and-fusion}(b). 
Then, we combine the sub-BN1' layer with the preceding CONV1 and
combine the sub-BN2' layer with both the ReLU layer and the following
CONV2 layer.
As opposed to the forward pass, the backward pass operations use not
only the partial derivatives on ifmaps/ofmaps but also the ifmaps generated
from the forward pass. 
Also, during the backward pass, CONV layers require twice as many
computations and memory accesses as those of the forward pass to
compute the partial derivatives on weight filters.
This increases memory accesses that cannot be reduced by the proposed
BNFF, making it less effective on the backward pass; but BNFF still
removes five memory sweeps per BN layer. 

BNFF can also be implemented across CPLs.
This Inter-Composite-Layer Fusion (ICF) requires fusing the first
sub-BN1 layer of a CPL after Fission with the corresponding
Concat (or Split) layer, not the CONV layer.
On the forward pass, because a Split layer is implemented as a pointer
passing in our reference implementation, a sub-BN1 layer is fused with
a Concat layer. 
On the backward pass, the sub-BN1' layer is fused with the corresponding
Split layer.
This completely removes all memory accesses on BN layers within DenseNet's
CPLs.
We implemented BNFF within CPLs and present experimental 
results measured from the latest chip processor in
Section~\ref{sec:evaluation}.
We estimate additional performance enhancement enabled by ICF, leaving
implementation for future work.

\vspace{-0.03in}
\section{Experimental Setup}
\label{sec:experimental-setup}
We quantify the performance improvement of the proposed BN Fission-n-Fusion
using DenseNet-121 and ResNet-50.
%
%
We used Caffe~\cite{arxiv-2014-caffe} (Intel Caffe 1.0.7, which was released
December 2017~\cite{intel-caffe}, for Intel processors) as a reference CNN framework.
We mainly used Intel's latest chip multiprocessor (Skylake-based Xeon Gold
6138~\cite{skylake}) and GPU (Nvidia Pascal Titan X~\cite{ieeemicro-2017-nvidia-pascal})  for performance evaluation.
The CPU system consists of 20 cores with 40 AVX-512 FMA (fused multiply-add)
units per socket, achieving a peak performance of 3.34TFLOPS with two sockets.
Two sockets have twelve memory channels, each with DDR4-2400 DRAM modules 
achieving up to 230.4GB/s.
%
To observe performance trends with a higher FLOP/B ratio, we also controlled
the data transfer rates of the memory channels.
%

%
\begin{figure}[!tb]
  \center
  \includegraphics[width=\columnwidth]{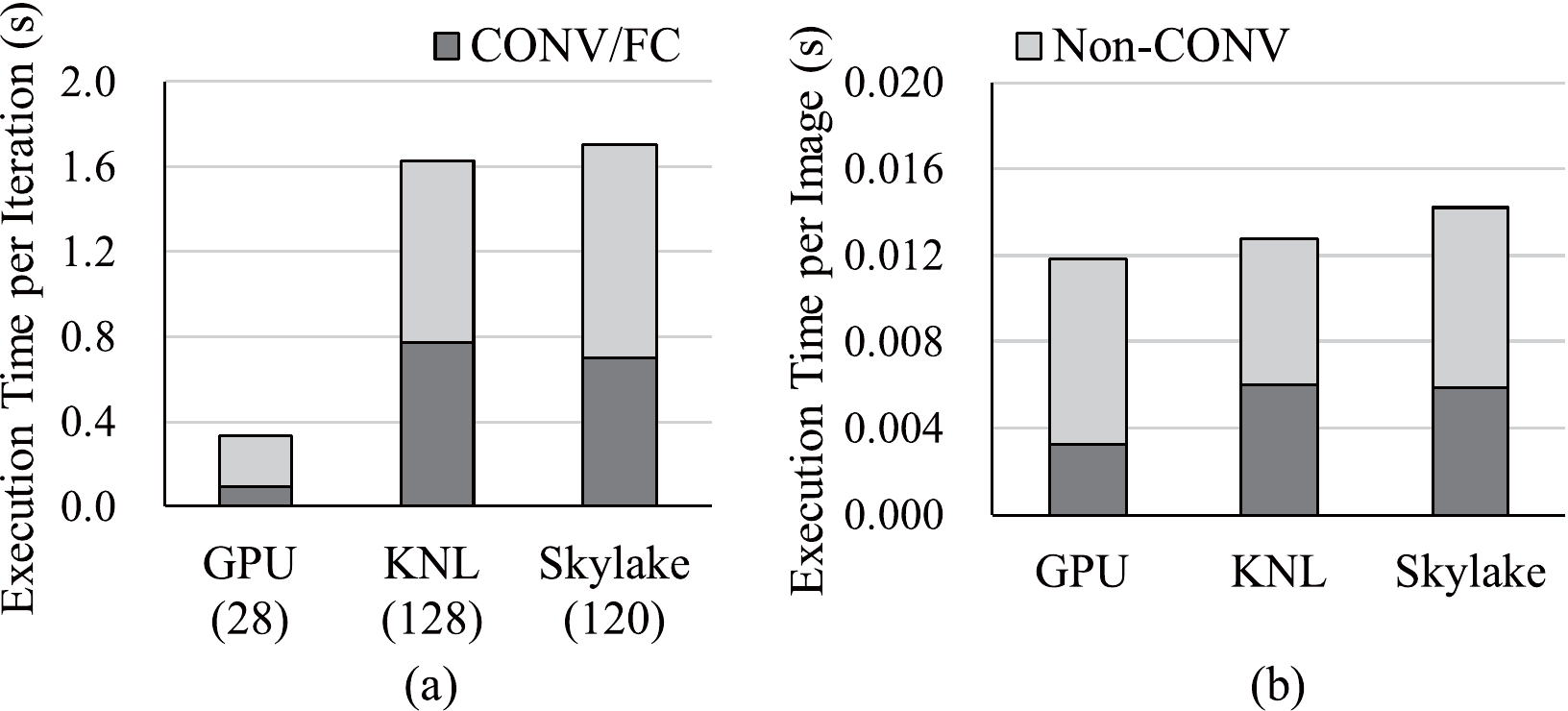}
  \caption{
    CONV/FC vs. non-CONV execution time of DenseNet-121 among 
    data-parallel architectures: GPU, KNL (Knights Landing), and 
    Skylake-based Xeon.
    (a) Execution time per iteration (the numbers in parentheses
    indicate mini-batch size),
    (b) Execution time per image (normalized by the mini-batch size.)
  }
  \label{fig:comparison-over-cnn-accelerators}
  \vspace{-0.05in}
\end{figure}

\begin{table}[!tb]
  \centering
  \caption{Peak single-precision floating point performance and peak memory bandwidth on the latest data-parallel architectures.}
  \begin{adjustbox}{width=\columnwidth}
    \begin{tabular}{@{}cccc@{}}
    \toprule
    \multicolumn{1}{l}{}   & \multicolumn{1}{c}{}     & \multicolumn{1}{c}{Main-memory} \\ 
    \multicolumn{1}{l}{Architectures}   & \multicolumn{1}{c}{TFLOPS}     & \multicolumn{1}{c}{BW (GB/s)} \\ 
    \midrule
      \multicolumn{1}{l}{Intel Xeon Skylake (2-socket)}  & \multicolumn{1}{c}{3.34}   & \multicolumn{1}{c}{230.4}      \\ [4pt]
      \multicolumn{1}{l}{Intel Xeon Phi Knights Landing}  & \multicolumn{1}{c}{5.30}   & \multicolumn{1}{c}{400.0}      \\ [4pt]
      \multicolumn{1}{l}{Nvidia GPU Pascal Titan X}      & \multicolumn{1}{c}{10.0}   & \multicolumn{1}{c}{480.0}      \\ 
    \bottomrule
    \end{tabular}
  \end{adjustbox}
  \label{tab:CNN-accelerators}
  \vspace{-0.05in}
\end{table}

We used Intel's Skylake-based Xeon processor because it supports a
highly-optimized open-source CNN library (MKL-DNN v0.11~\cite{mkldnn})
which can be modified for implementing BNFF.
It supports extracting a variety of detailed hardware statistics,
which can be helpful for analyzing BNFF benefits as well as for debugging.
%
%
%
Figure~\ref{fig:comparison-over-cnn-accelerators} compares the
execution time of DenseNet-121 over Pascal-based GPU
(Titan X) using cuDNN, Knights Landing Xeon Phi
(KNL)~\cite{ieeemicro-2016-knl}, and Skylake-based Xeon
(see Table~\ref{tab:CNN-accelerators} for a summary of their peak performance).
We set the mini-batch sizes of KNL and Skylake-based Xeon to 128 and 120,
respectively.  
However, due to memory capacity limitation, we set the mini-batch
size of the Pascal-based GPU to 28.

All three architectures spend more time on non-CONV layers compared
to CONV layers, demonstrating the importance of optimizing these
non-CONV layers (see Figure~\ref{fig:comparison-over-cnn-accelerators}(a)).
In Table~\ref{tab:CNN-accelerators}, Skylake-based Xeon has 1.6$\times$ and 3.0$\times$ 
lower peak performance than KNL and GPU.
However, execution time per image is similar to the others
(see Figure~\ref{fig:comparison-over-cnn-accelerators}(b))
because Skylake-based Xeon fully utilizes computing units on 
all CONV layers compared to the other architectures.


\begin{figure}[!tb]
  \center
  \subfloat{\includegraphics[width=\columnwidth]{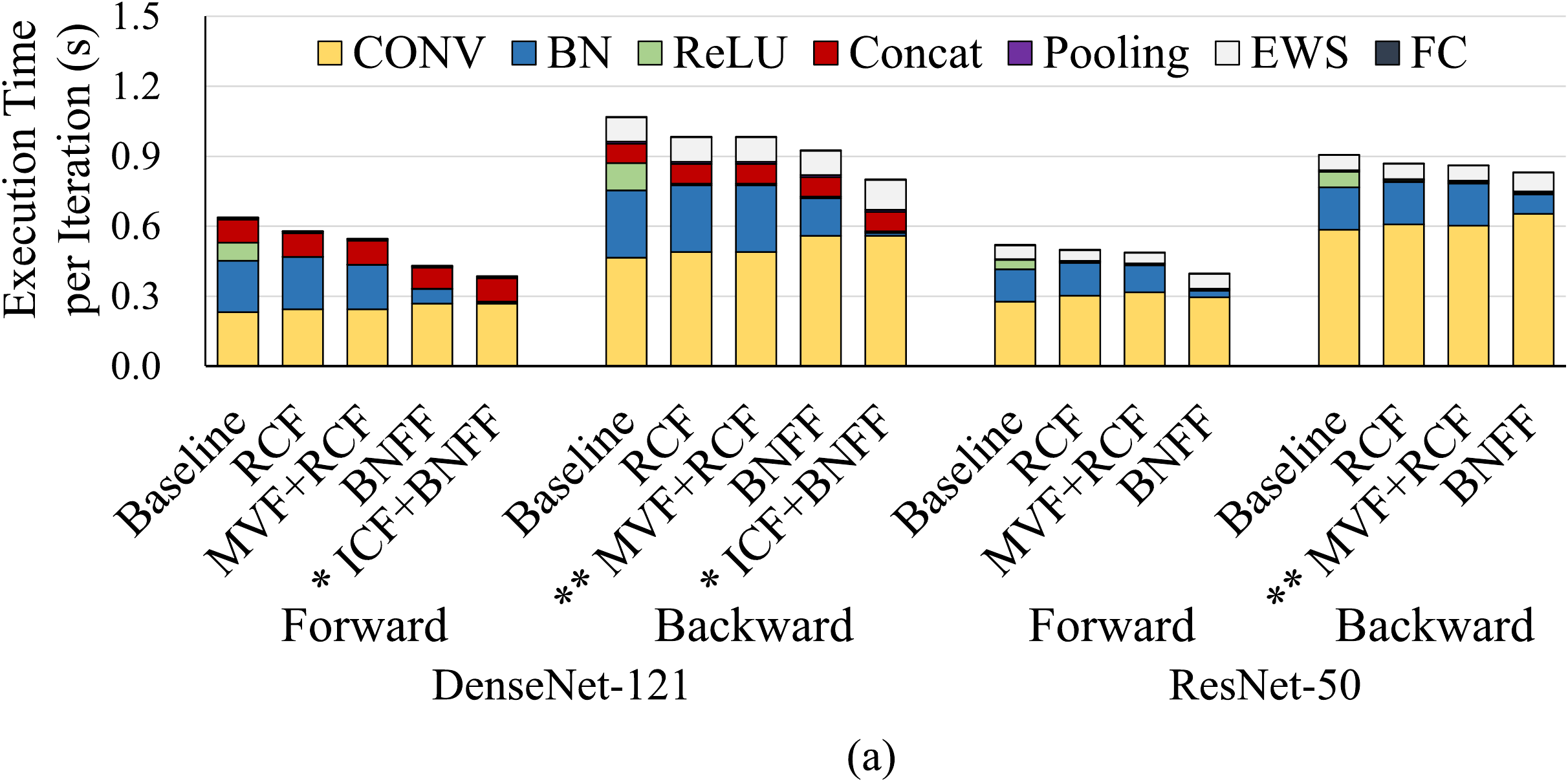}}
  \hspace{0.01in}
  \vspace{-0.05in}
  \subfloat{\includegraphics[width=\columnwidth]{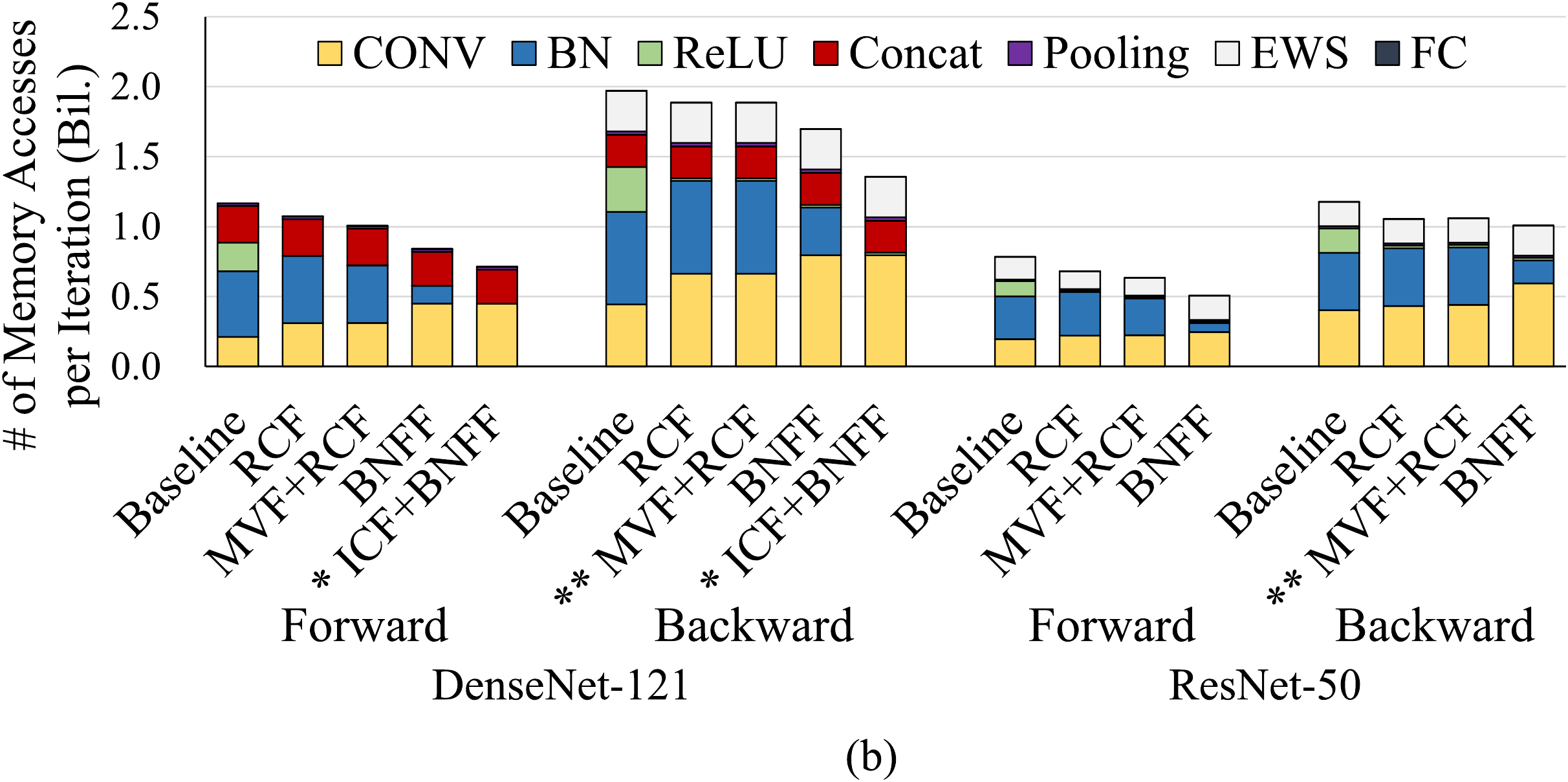}}
  \center
  \vspace{-0.15in}
  \caption{For DenseNet-121 and ResNet-50, (a) execution time and 
    (b) the number of memory accesses per iteration
    by applying RCF (ReLU-CONV Fusion), RCF+MVF (Mean/Variance Fusion), 
    BNFF (BN Fission-n-Fusion), and BNFF+ICF
    (Inter Composite-layer Fusion, DenseNet only). Mini-batch size is 120.
    *ICF improvement was estimated while the others were measured from Intel Xeon Skylake.
    %
    **MVF is not applicable to backward pass.
  }
  \label{fig:performance-memory}
\end{figure}

\section{Evaluation}
\label{sec:evaluation}
We provide the evaluation results of applying BN Fission-n-Fusion
on DenseNet-121 and ResNet-50 training.
To understand the quantitative improvements in detail, we have chosen four
different scenarios and have evaluated their execution time and the number of
memory accesses. 
The four scenarios are 
RCF (ReLU-CONV Fusion), 
MVF (Mean/Variance Fusion) together with RCF, 
BNFF (BN Fission-n-Fusion) that includes both MVF and RCF, 
and BNFF with ICF (Inter Composite-layer Fusion, applicable only to DenseNet).
While the performance improvement of RCF, MVF, and BNFF are from 
real-machine experiments, the improvement of ICF was estimated in line with 
BNFF improvement. 
The results are plotted in Figure~\ref{fig:performance-memory}.
As a quick summary, the gain of BNFF over the baseline turns out to be
47.9\% (30.8\%) for the forward pass and 15.4\% (9.0\%) for the backward pass, and 
overall 25.7\% (16.1\%) for the training process of DenseNet-121 
(ResNet-50) using the Xeon processor, respectively.

In the following, we discuss the results of the four different scenarios
using the Xeon processor with DenseNet-121, unless mentioned otherwise.
First, when only RCF is applied, the overall execution speed is increased by 9.2\%.
For the baseline, the number of memory accesses by ReLU layers
takes up 16.8\% of the total memory accesses.
By merging a ReLU layer with its adjacent CONV layer, the merged layer 
ends up with much less computation and memory access time compared to 
the total time needed for ReLU and CONV.
The gain is similar for the forward pass and the backward pass.
%
Second, when MVF is applied on top of RCF, the overall execution speed is 
increased by an additional 1.7\% compared to RCF.
Because MVF can be applied only to the forward pass,
the gain is entirely from 5.5\% of improvement in the
forward pass. 
Third, when BNFF is applied, where BNFF includes both MVF and RCF, 
the overall execution speed is improved by 25.7\% compared to the baseline
and 14.8\% compared to the MVF+RCF. 
By applying BNFF, memory access is reduced by 19.1\%.
Besides the reduced main-memory accesses, fewer subroutine calls and 
lower cache miss rates due to an improved cache pollution environment
by Fusion also contribute to the performance gains.
Lastly, when ICF is applied on top of BNFF, the overall execution speed is 
estimated to be increased by 43.7\% compared to the baseline.
With ICF, all the BN layers, including the ones that 
are on the boundaries of composite layers, benefit from Fission-n-Fusion.
Thus, an additional 18\% of improvement over BNFF is expected with ICF.


BNFF divides a BN layer into sub-BN1 and sub-BN2 layers and then fuses
them with the preceding and the following layers. 
Through this restructuring, the memory accesses of BN layers 
are entirely removed for the BN layers that do not cross the 
composite layer boundaries.  
As the result, the forward pass performance is improved by 47.9\%.
For the backward pass, however, the gain is only 15.4\%
because the CONV layers of backward pass need to perform 
twice as many computations and memory accesses compared to the 
forward pass. 
As CONV is heavier in computation for the backward pass, the relative portion
of BN is smaller and the gain of BNFF is smaller, too. 
We also observed that a Split layer in the backward pass requires
more frequent memory accesses compared to a Split layer in the forward pass.
For the forward pass, only the pointer needs to be passed.

\begin{figure}[!tb]
  \center
  \includegraphics[width=0.9\columnwidth]{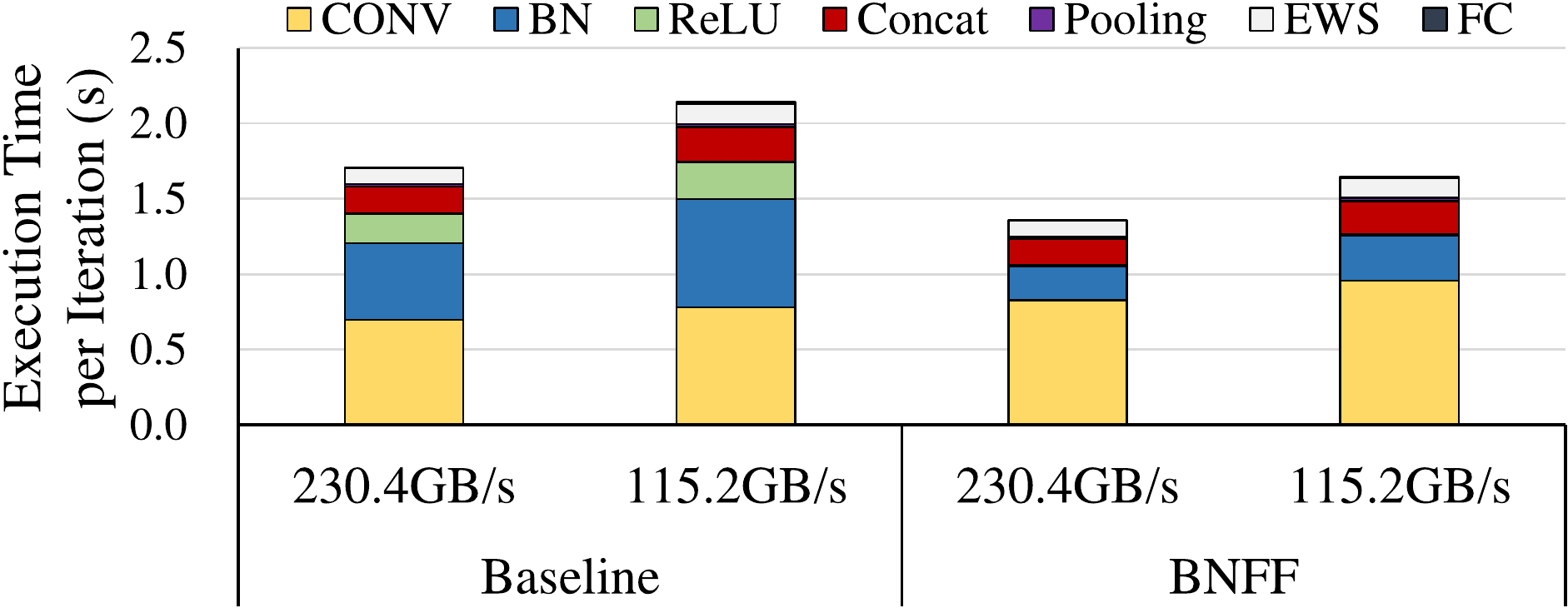}
  \vspace{-0.05in}
  \caption{Execution time comparison of baseline and BNFF for 230.4GB/s and 115.2GB/s memory bandwidth using the Xeon processor with DenseNet-121.}
  \label{fig:bw50-performance}
  \vspace{-0.15in}
\end{figure}

BNFF is more effective when the gap between computational power and 
main-memory bandwidth becomes wider (i.e., higher FLOP/B).
As an additional evaluation, we tried reducing the peak memory bandwidth to 
the half (115.2GB/s) of the original, 
and the evaluation results are shown in Figure~\ref{fig:bw50-performance}.
For the baseline, the portion of non-CONV layers in the total execution time 
increases from 58.9\% to 63.0\% as the peak memory bandwidth is reduced by half.
This is because the non-CONV layers are relatively more sensitive to 
memory bandwidth, whereas the CONV layers are more compute-intensive.
%
%
For BNFF, both non-CONV and CONV layers suffer from the reduced 
memory bandwidth.
Because many of the BN layers were optimized by BNFF already, 
it turns out that the degradation level is comparable for 
both CONV and non-CONV layers.
A large portion of the CONV layers in DenseNet-121
also demand high main-memory bandwidth because they have smaller weight
matrices (mostly 1$\times$1 and 3$\times$3) and hence their
performance is also influenced by the reduced main-memory bandwidth.
The gain of BNFF turns out to be 30.1\% for 115.2GB/s of peak main-memory bandwidth 
(c.f. 25.7\% for 230.4GB/s).

We also evaluated BNFF on GPU using Caffe.
We used Nvidia CUTLASS~\cite{nvidia-cutlass}, an open-source linear algebra library which enables us to implement BNFF,
because other highly-optimized but closed-source GPU libraries such as 
cuBLAS~\cite{nvidia-cublas} and cuDNN~\cite{arxiv-2014-cudnn} are difficult
to modify.\footnote{Compared to the cuDNN implementation used in Section~\ref{sec:experimental-setup},
the baseline CUTLASS implementation (with the mini-batch size of 16) is 3.6$\times$ slower.}
As for fusing convolution and mean-variance calculation for BN,
when each GPU thread stores one output value during convolution,
we store the output and its square value to shared memory, which
is shared among a thread block.
Prior to the end of the convolution operation, we perform 
per-thread-block reduction, followed by inter-thread-block 
reduction and normalization.
Comparing the baseline with the CUTLASS library,
the performance improvement by implementing RCF, MVF+RCF, and BNFF
on DenseNet-121 (ResNet-50)
is 0.7\% (0.3\%), 1.8\% (0.9\%), and 17.5\% (7.8\%), respectively.

\vspace{-0.07in}
\section{Related Work}
\label{sec:relatedwork}
\vspace{-0.07in}

There has been a large body of previous approaches to reduce memory accesses through increasing data reuse. 
We categorized the approaches into several groups, according to their purposes.

\textbf{Maximizing data reuse:} To reduce memory access cost, many proposed hardware accelerators have tried to maximize data reuse. 
DianNao~\cite{asplos-2014-diannao} optimized memory access to individual storage structures of weights and input/output feature maps, 
thereby alleviating inefficiencies in accessing off-chip memory. 
DaDianNao~\cite{micro-2014-dadiannao} 
also lowered off-chip memory accesses by putting all the weights on the on-chip memory. 
%
%
With regard to the on-chip data transfer, a dataflow scheme called row stationary enables near-optimal data reuse for the CONV 
layers~\cite{isca-2016-eyeriss}. 
%
%
Most of these proposals are limited to the inference process.

\textbf{Pruning and approximate computing:} By eliminating redundant parameters in weights and feature maps, or reducing their precision, 
their data sizes 
could be greatly reduced, leading to reduction in memory accesses. 
Deep compression~\cite{iclr-2015-deepcompression} reduces the model size of VGG-16 up to 49$\times$ by applying network pruning and quantization. 
EIE~\cite{isca-2016-eie} 
compresses the weights and the input feature maps.
%
CirCNN~\cite{micro-2017-circnn} uses 
circulant matrices to reduce computation and time complexity, with a minor accuracy drop. 
%
The aforementioned approaches mostly focused on the CONV/FC layers.

\textbf{Fusing and blending layers:} 
Jung et al.~\cite{cal-2017-partitioning} improves CNN inference performance by statistically spreading main-memory accesses 
through letting each compute unit operate asynchronously. 
Fused-layer~\cite{micro-2016-fused} is similar to our work as it fuses multiple CNN layers to exploit inter-layer data reusability. 
%
As opposed to Fused-layer, however, we target BN and training process, where tight dependency over a large 
volume of data makes fusing multiple CONV layers a daunting task.
Many machine learning frameworks, such as Caffe and TensorFlow~\cite{tensorflow}, have already implemented fusing multiple layers (i.e., CONV/BN with ReLU layers), targeting to optimize inter-layer data reuse. 
However, those implementations have limited potentials in that they do not fully fuse BN with other layer types, not considering the complex data dependency.
%

\textbf{Training acceleration:} The optimization strategies above mainly target the inference process. 
Several studies accelerate the training process~\cite{isca-2017-stochastic, micro-2017-circnn, micro-2016-vdnn, isca-2017-scaledeep}, but none of those focused on the non-CONV layers, taking a significant portion of 
training time of the latest CNN models.
Gist~\cite{isca-2018-gist} optimizes both CONV and non-CONV layers, although
it focuses on memory footprint reduction to enable using larger batch sizes.

\section{Conclusion}
\label{sec:conclusion}

The existing studies on CNN acceleration were mainly limited to convolutional (CONV) 
and fully-connected (FC) layers.
CNN models, however, are continuously evolving and the latest models can have hundreds of layers 
consisting of a variety of layer types. 
Among the non-CONV layers of the latest CNN models, we have found that batch normalization (BN) layers 
consume a significant portion of the execution time during training. 
A further analysis on BN's computation and memory access characteristics showed that the sequential 
and less computationally-intensive nature of the calculations with a strict dependency over a large
dataset causes an excessive memory access demand and makes fusing multiple CONV layers a daunting
task.
To address the issue, we have proposed a new type of CNN acceleration called Fission-n-Fusion, where 
BN is restructured to reduce memory access.
Experiments on a latest chip multiprocessor showed that the proposed BN restructuring 
can improve the training performance of DenseNet-121 by 47.9\% for forward pass 
and by 15.4\% for backward pass, leading to overall 25.7\% improvement.
Applying the BN restructuring to GPU with an open-source
linear algebra library also showed 17.4\% of performance improvement.
%
%
The large improvement suggests that non-CONV layers are important candidates for 
acceleration and that future research should pay keen attention to how CNN models 
evolve.

\section*{Acknowledgements}
This work was supported by Samsung Advanced Institute of
Technology, the Engineering Research Center Program 
through the NRF of Korea funded by the Korean Government MSIT (NRF-2018R1A5A1059921), 
and another NRF grant (NRF-2017R1E1A1A03070560).

\bibliography{paper}

\begin{thebibliography}{50}
\providecommand{\natexlab}[1]{#1}
\providecommand{\url}[1]{\texttt{#1}}
\expandafter\ifx\csname urlstyle\endcsname\relax
  \providecommand{\doi}[1]{doi: #1}\else
  \providecommand{\doi}{doi: \begingroup \urlstyle{rm}\Url}\fi

\bibitem[Abadi et~al.(2016)Abadi, Barham, Chen, Chen, Davis, Dean, Devin,
  Ghemawat, Irving, Isard, Kudlur, Levenberg, Monga, Moore, Murray, Steiner,
  Tucker, Vasudevan, Warden, Wicke, Yu, and Zheng]{tensorflow}
Abadi, M., Barham, P., Chen, J., Chen, Z., Davis, A., Dean, J., Devin, M.,
  Ghemawat, S., Irving, G., Isard, M., Kudlur, M., Levenberg, J., Monga, R.,
  Moore, S., Murray, D.~G., Steiner, B., Tucker, P.~A., Vasudevan, V., Warden,
  P., Wicke, M., Yu, Y., and Zheng, X.
\newblock {TensorFlow: {A} System for Large-Scale Machine Learning}.
\newblock In \emph{Proceedings of the 12th {USENIX} Symposium on Operating
  Systems Design and Implementation (OSDI)}, pp.\  265--283, 2016.

\bibitem[Albericio et~al.(2016)Albericio, Judd, Hetherington, Aamodt, Jerger,
  and Moshovos]{isca-2016-cnvlutin}
Albericio, J., Judd, P., Hetherington, T.~H., Aamodt, T.~M., Jerger, N. D.~E.,
  and Moshovos, A.
\newblock {Cnvlutin: Ineffectual-Neuron-Free Deep Neural Network Computing}.
\newblock In \emph{Proceedings of the 43rd Annual International Symposium on
  Computer Architecture (ISCA)}, pp.\  1--13, 2016.

\bibitem[Alwani et~al.(2016)Alwani, Chen, Ferdman, and
  Milder]{micro-2016-fused}
Alwani, M., Chen, H., Ferdman, M., and Milder, P.
\newblock {Fused-Layer {CNN} Accelerators}.
\newblock In \emph{Proceedings of the 49th Annual IEEE/ACM International
  Symposium on Microarchitecture (MICRO)}, pp.\  22:1--22:12, 2016.

\bibitem[Bergstra \& Bengio(2012)Bergstra and Bengio]{bergstra2012random}
Bergstra, J. and Bengio, Y.
\newblock {Random Search for Hyper-Parameter Optimization}.
\newblock \emph{Journal of Machine Learning Research}, 13:\penalty0 281--305,
  2012.

\bibitem[Bjorck et~al.(2018)Bjorck, Gomes, Selman, and
  Weinberger]{nips-2018-understandingbn}
Bjorck, N., Gomes, C.~P., Selman, B., and Weinberger, K.~Q.
\newblock {Understanding Batch Normalization}.
\newblock In \emph{Proceedings of the 31th Advances in Neural Information
  Processing Systems (NIPS)}, pp.\  7705--7716, 2018.

\bibitem[{Canziani} et~al.(2016){Canziani}, {Paszke}, and
  {Culurciello}]{arxiv-2016-dnn-analysis}
{Canziani}, A., {Paszke}, A., and {Culurciello}, E.
\newblock {An Analysis of Deep Neural Network Models for Practical
  Applications}.
\newblock \emph{ArXiv e-prints}, May 2016.

\bibitem[Chen et~al.(2014{\natexlab{a}})Chen, Du, Sun, Wang, Wu, Chen, and
  Temam]{asplos-2014-diannao}
Chen, T., Du, Z., Sun, N., Wang, J., Wu, C., Chen, Y., and Temam, O.
\newblock {DianNao: A Small-Footprint High-Throughput Accelerator for
  Ubiquitous Machine-Learning}.
\newblock In \emph{Proceedings of the 19th International Conference on
  Architectural Support for Programming Languages and Operating Systems
  (ASPLOS)}, pp.\  269--284, 2014{\natexlab{a}}.

\bibitem[Chen et~al.(2016{\natexlab{a}})Chen, Duan, Houthooft, Schulman,
  Sutskever, and Abbeel]{chen2016infogan}
Chen, X., Duan, Y., Houthooft, R., Schulman, J., Sutskever, I., and Abbeel, P.
\newblock {Infogan: Interpretable Representation Learning by Information
  Maximizing Generative Adversarial Nets}.
\newblock In \emph{Proceedings of the 29th Advances in Neural Information
  Processing Systems (NIPS)}, pp.\  2172--2180, 2016{\natexlab{a}}.

\bibitem[Chen et~al.(2014{\natexlab{b}})Chen, Luo, Liu, Zhang, He, Wang, Li,
  Chen, Xu, Sun, and Temam]{micro-2014-dadiannao}
Chen, Y., Luo, T., Liu, S., Zhang, S., He, L., Wang, J., Li, L., Chen, T., Xu,
  Z., Sun, N., and Temam, O.
\newblock {DaDianNao: A Machine-Learning Supercomputer}.
\newblock In \emph{Proceedings of the 47th Annual IEEE/ACM International
  Symposium on Microarchitecture (MICRO)}, pp.\  609--622, 2014{\natexlab{b}}.

\bibitem[Chen et~al.(2016{\natexlab{b}})Chen, Emer, and Sze]{isca-2016-eyeriss}
Chen, Y., Emer, J., and Sze, V.
\newblock {Eyeriss: A Spatial Architecture for Energy-Efficient Dataflow for
  Convolutional Neural Networks}.
\newblock In \emph{Proceedings of the 42nd Annual International Symposium on
  Computer Architecture (ISCA)}, pp.\  367--379, 2016{\natexlab{b}}.

\bibitem[Chetlur et~al.(2014)Chetlur, Woolley, Vandermersch, Cohen, Tran,
  Catanzaro, and Shelhamer]{arxiv-2014-cudnn}
Chetlur, S., Woolley, C., Vandermersch, P., Cohen, J., Tran, J., Catanzaro, B.,
  and Shelhamer, E.
\newblock {cuDNN: Efficient Primitives for Deep Learning}.
\newblock \emph{ArXiv e-prints}, October 2014.

\bibitem[Chi et~al.(2016)Chi, Li, Xu, Zhang, Zhao, Liu, Wang, and
  Xie]{isca-2016-prime}
Chi, P., Li, S., Xu, C., Zhang, T., Zhao, J., Liu, Y., Wang, Y., and Xie, Y.
\newblock {PRIME: A Novel Processing-in-Memory Architecture for Neural Network
  Computation in ReRAM-Based Main Memory}.
\newblock In \emph{Proceedings of the 42nd Annual International Symposium on
  Computer Architecture (ISCA)}, pp.\  27--39, 2016.

\bibitem[Cogswell et~al.(2015)Cogswell, Ahmed, Girshick, Zitnick, and
  Batra]{cogswell2015reducing}
Cogswell, M., Ahmed, F., Girshick, R.~B., Zitnick, L., and Batra, D.
\newblock {Reducing Overfitting in Deep Networks by Decorrelating
  Representations}.
\newblock \emph{ArXiv e-prints}, November 2015.

\bibitem[De~Sa et~al.(2017)De~Sa, Feldman, R{\'e}, and
  Olukotun]{isca-2017-stochastic}
De~Sa, C., Feldman, M., R{\'e}, C., and Olukotun, K.
\newblock {Understanding and Optimizing Asynchronous Low-Precision Stochastic
  Gradient Descent}.
\newblock In \emph{Proceedings of the 44th Annual International Symposium on
  Computer Architecture (ISCA)}, pp.\  561--574, 2017.

\bibitem[Ding et~al.(2017)Ding, Liao, Wang, Li, Liu, Zhuo, Wang, Qian, Bai,
  Yuan, Ma, Zhang, Tang, Qiu, Lin, and Yuan]{micro-2017-circnn}
Ding, C., Liao, S., Wang, Y., Li, Z., Liu, N., Zhuo, Y., Wang, C., Qian, X.,
  Bai, Y., Yuan, G., Ma, X., Zhang, Y., Tang, J., Qiu, Q., Lin, X., and Yuan,
  B.
\newblock {CirCNN: Accelerating and Compressing Deep Neural Networks Using
  Block-circulant Weight Matrices}.
\newblock In \emph{Proceedings of the 50th Annual IEEE/ACM International
  Symposium on Microarchitecture (MICRO)}, pp.\  395--408, 2017.

\bibitem[Doweck et~al.(2017)Doweck, Kao, y.~Lu, Mandelblat, Rahatekar,
  Rappoport, Rotem, Yasin, and Yoaz]{skylake}
Doweck, J., Kao, W.~F., y.~Lu, A.~K., Mandelblat, J., Rahatekar, A., Rappoport,
  L., Rotem, E., Yasin, A., and Yoaz, A.
\newblock {Inside 6th-Generation Intel Core: New Microarchitecture Code-Named
  Skylake}.
\newblock \emph{Micro, IEEE}, 37\penalty0 (2):\penalty0 52--62, 2017.

\bibitem[Du et~al.(2015)Du, Fasthuber, Chen, Ienne, Li, Luo, Feng, Chen, and
  Temam]{isca-2015-shidiannao}
Du, Z., Fasthuber, R., Chen, T., Ienne, P., Li, L., Luo, T., Feng, X., Chen,
  Y., and Temam, O.
\newblock {ShiDianNao: Shifting Vision Processing Closer to the Sensor}.
\newblock In \emph{Proceedings of the 42nd Annual International Symposium on
  Computer Architecture (ISCA)}, pp.\  92--104, 2015.

\bibitem[Foley \& Danskin(2017)Foley and Danskin]{ieeemicro-2017-nvidia-pascal}
Foley, D. and Danskin, J.
\newblock {Ultra-Performance Pascal GPU and NVLink Interconnect}.
\newblock \emph{Micro, IEEE}, 37\penalty0 (2):\penalty0 7--17, 2017.

\bibitem[Goyal et~al.(2017)Goyal, Doll{\'{a}}r, Girshick, Noordhuis,
  Wesolowski, Kyrola, Tulloch, Jia, and He]{arxiv-2017-facebook}
Goyal, P., Doll{\'{a}}r, P., Girshick, R.~B., Noordhuis, P., Wesolowski, L.,
  Kyrola, A., Tulloch, A., Jia, Y., and He, K.
\newblock {Accurate, Large Minibatch {SGD:} Training ImageNet in 1 Hour}.
\newblock \emph{ArXiv e-prints}, June 2017.

\bibitem[Han et~al.(2015{\natexlab{a}})Han, Mao, and
  Dally]{iclr-2015-deepcompression}
Han, S., Mao, H., and Dally, W.~J.
\newblock {Deep Compression: Compressing Deep Neural Networks with Pruning,
  Trained Quantization and Huffman Coding}.
\newblock \emph{ArXiv e-prints}, October 2015{\natexlab{a}}.

\bibitem[Han et~al.(2015{\natexlab{b}})Han, Pool, Tran, and
  Dally]{nips-2015-prunning}
Han, S., Pool, J., Tran, J., and Dally, W.~J.
\newblock {Learning Both Weights and Connections for Efficient Neural Network}.
\newblock In \emph{Proceedings of the 28th International Conference on Neural
  Information Processing Systems (NIPS)}, pp.\  1135--1143, 2015{\natexlab{b}}.

\bibitem[Han et~al.(2016)Han, Liu, Mao, Pu, Pedram, Horowitz, and
  Dally]{isca-2016-eie}
Han, S., Liu, X., Mao, H., Pu, J., Pedram, A., Horowitz, M.~A., and Dally,
  W.~J.
\newblock {EIE: Efficient Inference Engine on Compressed Deep Neural Network}.
\newblock In \emph{Proceedings of the 42nd Annual International Symposium on
  Computer Architecture (ISCA)}, pp.\  243--254, 2016.

\bibitem[He et~al.(2016{\natexlab{a}})He, Zhang, Ren, and
  Sun]{arxiv-2016-identity}
He, K., Zhang, X., Ren, S., and Sun, J.
\newblock {Identity Mappings in Deep Residual Networks}.
\newblock \emph{ArXiv e-prints}, March 2016{\natexlab{a}}.

\bibitem[He et~al.(2016{\natexlab{b}})He, Zhang, Ren, and
  Sun]{cvpr-2016-resnet}
He, K., Zhang, X., Ren, S., and Sun, J.
\newblock {Deep Residual Learning for Image Recognition}.
\newblock In \emph{Proceedings of the IEEE Conference on Computer Vision and
  Pattern Recognition (CVPR)}, pp.\  770--778, 2016{\natexlab{b}}.

\bibitem[Howard et~al.(2017)Howard, Zhu, Chen, Kalenichenko, Wang, Weyand,
  Andreetto, and Adam]{arxiv-2017-mobilenet}
Howard, A.~G., Zhu, M., Chen, B., Kalenichenko, D., Wang, W., Weyand, T.,
  Andreetto, M., and Adam, H.
\newblock {MobileNets: Efficient Convolutional Neural Networks for Mobile
  Vision Applications}.
\newblock \emph{ArXiv e-prints}, April 2017.

\bibitem[Huang et~al.(2017)Huang, Liu, van~der Maaten, and
  Weinberger]{cvpr-2017-densenet}
Huang, G., Liu, Z., van~der Maaten, L., and Weinberger, K.~Q.
\newblock {Densely Connected Convolutional Networks}.
\newblock In \emph{Proceedings of the IEEE Conference on Computer Vision and
  Pattern Recognition (CVPR)}, pp.\  2261--2269, 2017.

\bibitem[Intel(2016)]{mkldnn}
Intel.
\newblock {Intel(R) Math Kernel Library for Deep Neural Networks}, 2016.
\newblock URL \url{https://github.com/01org/mkl-dnn}.

\bibitem[Intel(2017)]{intel-caffe}
Intel.
\newblock {Intel(R) Distribution of Caffe}, December 2017.
\newblock URL \url{https://github.com/intel/caffe}.

\bibitem[Ioffe \& Szegedy(2015)Ioffe and Szegedy]{pmlr-2015-bn}
Ioffe, S. and Szegedy, C.
\newblock {Batch Normalization: Accelerating Deep Network Training by Reducing
  Internal Covariate Shift}.
\newblock In \emph{Proceedings of the 32nd International Conference on Machine
  Learning (ICML)}, pp.\  448--456, 2015.

\bibitem[{Jain} et~al.(2018){Jain}, {Phanishayee}, {Mars}, {Tang}, and
  {Pekhimenko}]{isca-2018-gist}
{Jain}, A., {Phanishayee}, A., {Mars}, J., {Tang}, L., and {Pekhimenko}, G.
\newblock {Gist: Efficient Data Encoding for Deep Neural Network Training}.
\newblock In \emph{Proceedings of the 45th Annual International Symposium on
  Computer Architecture (ISCA)}, pp.\  776--789, 2018.

\bibitem[Jia et~al.(2014)Jia, Shelhamer, Donahue, Karayev, Long, Girshick,
  Guadarrama, and Darrell]{arxiv-2014-caffe}
Jia, Y., Shelhamer, E., Donahue, J., Karayev, S., Long, J., Girshick, R.~B.,
  Guadarrama, S., and Darrell, T.
\newblock {Caffe: Convolutional Architecture for Fast Feature Embedding}.
\newblock \emph{ArXiv e-prints}, June 2014.

\bibitem[Jouppi et~al.(2017)Jouppi, Young, Patil, Patterson, Agrawal, Bajwa,
  Bates, Bhatia, Boden, Borchers, Boyle, Cantin, Chao, Clark, Coriell, Daley,
  Dau, Dean, Gelb, Ghaemmaghami, Gottipati, Gulland, Hagmann, Ho, Hogberg, Hu,
  Hundt, Hurt, Ibarz, Jaffey, Jaworski, Kaplan, Khaitan, Killebrew, Koch,
  Kumar, Lacy, Laudon, Law, Le, Leary, Liu, Lucke, Lundin, MacKean, Maggiore,
  Mahony, Miller, Nagarajan, Narayanaswami, Ni, Nix, Norrie, Omernick,
  Penukonda, Phelps, Ross, Ross, Salek, Samadiani, Severn, Sizikov, Snelham,
  Souter, Steinberg, Swing, Tan, Thorson, Tian, Toma, Tuttle, Vasudevan,
  Walter, Wang, Wilcox, and Yoon]{isca-2017-tpu}
Jouppi, N.~P., Young, C., Patil, N., Patterson, D., Agrawal, G., Bajwa, R.,
  Bates, S., Bhatia, S., Boden, N., Borchers, A., Boyle, R., Cantin, P.-l.,
  Chao, C., Clark, C., Coriell, J., Daley, M., Dau, M., Dean, J., Gelb, B.,
  Ghaemmaghami, T.~V., Gottipati, R., Gulland, W., Hagmann, R., Ho, C.~R.,
  Hogberg, D., Hu, J., Hundt, R., Hurt, D., Ibarz, J., Jaffey, A., Jaworski,
  A., Kaplan, A., Khaitan, H., Killebrew, D., Koch, A., Kumar, N., Lacy, S.,
  Laudon, J., Law, J., Le, D., Leary, C., Liu, Z., Lucke, K., Lundin, A.,
  MacKean, G., Maggiore, A., Mahony, M., Miller, K., Nagarajan, R.,
  Narayanaswami, R., Ni, R., Nix, K., Norrie, T., Omernick, M., Penukonda, N.,
  Phelps, A., Ross, J., Ross, M., Salek, A., Samadiani, E., Severn, C.,
  Sizikov, G., Snelham, M., Souter, J., Steinberg, D., Swing, A., Tan, M.,
  Thorson, G., Tian, B., Toma, H., Tuttle, E., Vasudevan, V., Walter, R., Wang,
  W., Wilcox, E., and Yoon, D.~H.
\newblock {In-datacenter Performance Analysis of a Tensor Processing Unit}.
\newblock In \emph{Proceedings of the 44th Annual International Symposium on
  Computer Architecture (ISCA)}, pp.\  1--12, 2017.

\bibitem[Jung et~al.(2018)Jung, Lee, Rhee, and Ahn]{cal-2017-partitioning}
Jung, D., Lee, S., Rhee, W., and Ahn, J.
\newblock {Partitioning Compute Units in CNN Acceleration for Statistical
  Memory Traffic Shaping}.
\newblock \emph{IEEE Computer Architecture Letters}, 17\penalty0 (1):\penalty0
  72--75, 2018.

\bibitem[Kerr et~al.(2017)Kerr, Merrill, Demouth, and Tran]{nvidia-cutlass}
Kerr, A., Merrill, D., Demouth, J., and Tran, J.
\newblock {CUTLASS: Fast Linear Algebra in CUDA C++}, December 2017.
\newblock URL \url{https://devblogs.nvidia.com/cutlass-linear-algebra-cuda}.

\bibitem[Kim et~al.(2016)Kim, Kung, Chai, Yalamanchili, and
  Mukhopadhyay]{isca-2016-neurocube}
Kim, D., Kung, J., Chai, S., Yalamanchili, S., and Mukhopadhyay, S.
\newblock {Neurocube: A Programmable Digital Neuromorphic Architecture with
  High-Density 3D Memory}.
\newblock In \emph{Proceedings of the 42nd Annual International Symposium on
  Computer Architecture (ISCA)}, pp.\  380--392, 2016.

\bibitem[Krizhevsky et~al.(2012)Krizhevsky, Sutskever, and
  Hinton]{nips-2012-alexnet}
Krizhevsky, A., Sutskever, I., and Hinton, G.~E.
\newblock {ImageNet Classification with Deep Convolutional Neural Networks}.
\newblock In \emph{Proceedings of the 25th International Conference on Neural
  Information Processing Systems (NIPS)}, pp.\  1106--1114, 2012.

\bibitem[NVIDIA()]{nvidia-cublas}
NVIDIA.
\newblock {NVIDIA cuBLAS Library}.
\newblock URL \url{https://developer.nvidia.com/cublas}.

\bibitem[Rhu et~al.(2016)Rhu, Gimelshein, Clemons, Zulfiqar, and
  W.~Keckler]{micro-2016-vdnn}
Rhu, M., Gimelshein, N., Clemons, J., Zulfiqar, A., and W.~Keckler, S.
\newblock {vDNN: Virtualized Deep Neural Networks for Scalable,
  Memory-Efficient Neural Network Design}.
\newblock In \emph{Proceedings of the 49th Annual IEEE/ACM International
  Symposium on Microarchitecture (MICRO)}, pp.\  18:1--18:13, 2016.

\bibitem[Santurkar et~al.(2018)Santurkar, Tsipras, Ilyas, and
  Madry]{nips-2018-bnhelpoptimize}
Santurkar, S., Tsipras, D., Ilyas, A., and Madry, A.
\newblock {How Does Batch Normalization Help Optimization?}
\newblock In \emph{Proceedings of the 31th Advances in Neural Information
  Processing Systems (NIPS)}, pp.\  2488--2498, 2018.

\bibitem[Shafiee et~al.(2016)Shafiee, Nag, Muralimanohar, and
  Balasubramonian]{isca-2016-isaac}
Shafiee, A., Nag, A., Muralimanohar, N., and Balasubramonian, R.
\newblock {ISAAC: A Convolutional Neural Network Accelerator with In-Situ
  Analog Arithmetic in Crossbars}.
\newblock In \emph{Proceedings of the 42nd Annual International Symposium on
  Computer Architecture (ISCA)}, pp.\  14--26, 2016.

\bibitem[Silver et~al.(2016)Silver, Huang, Maddison, Guez, Sifre, van~den
  Driessche, Schrittwieser, Antonoglou, Panneershelvam, Lanctot, Dieleman,
  Grewe, Nham, Kalchbrenner, Sutskever, Lillicrap, Leach, Kavukcuoglu, Graepel,
  and Hassabis]{nature-2016-alphago}
Silver, D., Huang, A., Maddison, C., Guez, A., Sifre, L., van~den Driessche,
  G., Schrittwieser, J., Antonoglou, I., Panneershelvam, V., Lanctot, M.,
  Dieleman, S., Grewe, D., Nham, J., Kalchbrenner, N., Sutskever, I.,
  Lillicrap, T., Leach, M., Kavukcuoglu, K., Graepel, T., and Hassabis, D.
\newblock {Mastering the Game of Go with Deep Neural Networks and Tree Search}.
\newblock \emph{Nature}, 529\penalty0 (7587):\penalty0 484--489, 2016.

\bibitem[Simonyan \& Zisserman(2014)Simonyan and Zisserman]{arxiv-2014-vggnet}
Simonyan, K. and Zisserman, A.
\newblock {Very Deep Convolutional Networks for Large-Scale Image Recognition}.
\newblock \emph{ArXiv e-prints}, September 2014.

\bibitem[Snoek et~al.(2012)Snoek, Larochelle, and Adams]{snoek2012practical}
Snoek, J., Larochelle, H., and Adams, R.~P.
\newblock {Practical Bayesian Optimization of Machine Learning Algorithms}.
\newblock In \emph{Proceedings of the 25th Advances in Neural Information
  Processing Systems (NIPS)}, pp.\  2960--2968, 2012.

\bibitem[Sodani et~al.(2016)Sodani, Gramunt, Corbal, Kim, Vinod, Chinthamani,
  Hutsell, Agarwal, and Liu]{ieeemicro-2016-knl}
Sodani, A., Gramunt, R., Corbal, J., Kim, H.-S., Vinod, K., Chinthamani, S.,
  Hutsell, S., Agarwal, R., and Liu, Y.-C.
\newblock {Knights Landing: Second Generation Intel Xeon Phi Product}.
\newblock \emph{Micro, IEEE}, 36\penalty0 (2):\penalty0 34--46, 2016.

\bibitem[Sriniva \& Babu(2015)Sriniva and Babu]{arxiv-2015-datafree}
Sriniva, S. and Babu, R.~V.
\newblock {Data-free Parameter Pruning for Deep Neural Networks}.
\newblock \emph{ArXiv e-prints}, July 2015.

\bibitem[Szegedy et~al.(2016)Szegedy, Ioffe, Vanhoucke, and
  Alemi]{arxiv-2016-inception-v4}
Szegedy, C., Ioffe, S., Vanhoucke, V., and Alemi, A.
\newblock {Inception-v4, Inception-ResNet and the Impact of Residual
  Connections on Learning}.
\newblock \emph{ArXiv e-prints}, February 2016.

\bibitem[Venkataramani et~al.(2017)Venkataramani, Ranjan, Banerjee, Das,
  Avancha, Jagannathan, Durg, Nagaraj, Kaul, Dubey, and
  Raghunathan]{isca-2017-scaledeep}
Venkataramani, S., Ranjan, A., Banerjee, S., Das, D., Avancha, S., Jagannathan,
  A., Durg, A., Nagaraj, D., Kaul, B., Dubey, P., and Raghunathan, A.
\newblock {ScaleDeep: A Scalable Compute Architecture for Learning and
  Evaluating Deep Networks}.
\newblock In \emph{Proceedings of the 44th Annual International Symposium on
  Computer Architecture (ISCA)}, pp.\  13--26, 2017.

\bibitem[Xie et~al.(2016)Xie, Girshick, Doll{\'{a}}r, Tu, and
  He]{arxiv-2016-resnext}
Xie, S., Girshick, R.~B., Doll{\'{a}}r, P., Tu, Z., and He, K.
\newblock {Aggregated Residual Transformations for Deep Neural Networks}.
\newblock \emph{ArXiv e-prints}, November 2016.

\bibitem[Yang(2017)]{densenet-github}
Yang, S.
\newblock {densenet-github}, 2017.
\newblock URL \url{https://github.com/shicai/DenseNet-Caffe}.

\bibitem[Yang et~al.(2016)Yang, Pu, Rister, Bhagdikar, Richardson, Kvatinsky,
  Ragan{-}Kelley, Pedram, and Horowitz]{arxiv-2016-systematic}
Yang, X., Pu, J., Rister, B.~B., Bhagdikar, N., Richardson, S., Kvatinsky, S.,
  Ragan{-}Kelley, J., Pedram, A., and Horowitz, M.
\newblock {A Systematic Approach to Blocking Convolutional Neural Networks}.
\newblock \emph{ArXiv e-prints}, June 2016.

\end{thebibliography}
\bibliographystyle{sysml2019}

\end{document}